\documentclass[journal]{IEEEtran}
\ifCLASSINFOpdf
\else
\fi

\usepackage{cite}
\usepackage{amsmath,amssymb,amsfonts}
\usepackage{algorithmic}
\usepackage{graphicx}
\usepackage{textcomp}
\usepackage{xcolor}
\usepackage{algorithm}
\usepackage{multirow}
\usepackage{float}
\usepackage{caption}
\usepackage{subcaption}
\usepackage{bbm}
\usepackage{array}
\usepackage{dblfloatfix}
\usepackage{tabularx}
\usepackage{xcolor}
\usepackage{comment}
\begin{document}
%
\title{Anomaly Detection with Ensemble of Encoder and Decoder}
%
%
%

\author{ \IEEEauthorblockN{Xijuan Sun\IEEEauthorrefmark{1}, Di Wu\IEEEauthorrefmark{1}, Arnaud Zinflou\IEEEauthorrefmark{2},  Benoit Boulet\IEEEauthorrefmark{1}}\\
\IEEEauthorblockA{\IEEEauthorrefmark{1}Department of Electrical and Computer Engineering, McGill University, Montreal, Canada \\
xijuan.sun@mail.mcgill.ca; di.wu5@mcgill.ca; benoit.boulet@mcgill.ca}\\ \IEEEauthorblockA{\IEEEauthorrefmark{2}Hydro-Québec’s research institute, Montreal, Canada \\
zinflou.arnaud@hydroquebec.com}
\thanks{Di Wu is the corresponding author for this paper.}
}

%
%

\markboth{Journal of \LaTeX\ Class Files,~Vol.~14, No.~8, August~2023}%
{Shell \MakeLowercase{\textit{et al.}}: Bare Demo of IEEEtran.cls for IEEE Journals}
%



\maketitle

\begin{abstract}
Hacking and false data injection from adversaries can threaten power grids' everyday operations and cause significant economic loss. Anomaly detection in power grids aims to detect and discriminate anomalies caused by cyber attacks against the power system, which is essential for keeping power grids working correctly and efficiently. Different methods have been applied for anomaly detection, such as statistical methods and machine learning-based methods. Usually, machine learning-based methods need to model the normal data distribution. In this work, we propose a novel anomaly detection method by modeling the data distribution of normal samples via multiple encoders and decoders. Specifically, the proposed method maps input samples into a latent space and then reconstructs output samples from latent vectors. The extra encoder finally maps reconstructed samples to latent representations. During the training phase, we optimize parameters by minimizing the reconstruction loss and encoding loss. Training samples are re-weighted to focus more on missed correlations between features of normal data. Furthermore, we employ the long short-term memory model as encoders and decoders to test its effectiveness. We also investigate a meta-learning-based framework for hyper-parameter tuning of our approach. Experiment results on network intrusion and power system datasets demonstrate the effectiveness of our proposed method, where our models consistently outperform all baselines.
\end{abstract}


%
\IEEEpeerreviewmaketitle

\section{Introduction}
Smart grids involve technologies and monitors such as advanced meters, real-time controllers, and renewable energy resources~\cite{yadav2016review}. With the ever-increasing application of electrical appliances, smart grid systems have become indispensable to urban life~\cite{adiban2021step, wu2017two, wu2018machine}. 
However, various network intrusions and cyber attacks have seriously threatened power grids~\cite{ADIBAN2020101105, havlena2022accurate, tong2022fast}, such as data alteration and false data injection~\cite{zhangde2021tecting, mohammad2022anomaly}. 
Attacks on smart grids can cause abnormal operations and lead to the instabilities of power grids~\cite{adiban2018, zhang2021time, pan2022dynamic}. Therefore, detecting anomalies on power grids is of particular importance. In~\cite{bho2019on}, anomaly detection is formulated as a supervised learning problem. Several traditional supervised learning algorithms, such as random forest~\cite{lin2020anomaly} and K-nearest neighbor~\cite{gharaei2022}, have been utilized for anomaly detection. Nevertheless, collecting a large number of labeled cyber attacks on power grids is only sometimes feasible since it can be time-consuming and even impractical for specific scenarios. 
Also, various new attacks may arise with the advance of cyber attacks and intrusions. Models based on supervised learning may not work well against these unseen attacks. Due to the above reasons, treating anomaly detection on power grids as an unsupervised or semi-supervised learning problem is more effective~\cite{zenati2018}. Besides anomaly detection, machine learning-based methods have also been applied for other smart grid applications, such as electric load forecasting~\cite{wu2019multiple, wu2017boosting, lin2021spatial, lin2021residential, wu2022efficient, fu2022reinforcement} and electric vehicle charging scheduling~\cite{dang2019q, wu2018optimizing, wu2014neighborhood, dang2020ev, huang2020ensemble, dang2019advanced, xiong2015impact}.

In recent years, Generative Adversarial Network (GAN)~\cite{Goodfellow2014generative} have gained significant attention for different kinds of tasks, including anomaly detection tasks in~\cite{jeong2021poster, jiang2020discriminative}. GAN is a kind of generative model and has been widely used in various tasks, including natural language process~\cite{karuna2022generative}, image data augmentation~\cite{mudavathu2018aux}, and video prediction~\cite{hu2020generative}. 
GAN consists of a generator and a discriminator, capable of learning complex and high-dimensional data distribution via adversarial training.
In~\cite{akcay2019ganomaly}, an encoder-decoder-encoder network is used as the generator of the GAN model for image anomaly detection. However, GANs suffer from mode collapse~\cite{Bhag2020study}, and the performance of a single GAN can be unstable on anomaly detection tasks~\cite{koda2017on}. In~\cite{tol2017adagan}, the GAN model is combined with the AdaBoost algorithm for generating similar data distribution of input samples. In~\cite{Han_Chen_Liu_2021}, an algorithm is proposed using the ensemble learning framework on GANs for anomaly detection.  


In practice, no single machine learning algorithm can be well adapted to all datasets in a domain. Choosing efficient algorithms for different tasks is always tricky and time-consuming~\cite{maher2019smart}, and hyper-parameter tuning for machine learning-based methods is also an intractable problem~\cite{maher2019smart}. In most real-world cases, experts are employed to tune the hyper-parameters cautiously to obtain a robust model~\cite{molina2012meta}. Such an iterative and explorative hyper-parameter tuning process is usually tedious, and hiring experts is expensive and only sometimes possible in the real world~\cite{maher2019smart}. Thus, there is a rise in exploring algorithms for automatic model selection and hyper-parameter tuning~\cite{maher2019smart}. In recent years, meta-learning has been investigated for automatic model selection and hyper-parameter tuning~\cite{maher2019smart}. The meta-learning approach employs a concept called "learning to learn," which assists models in adapting more efficiently to new tasks~\cite{so2021exploring}. Several models are trained on different tasks in the meta-learning framework to obtain the meta-knowledge utilized for adapting new and unknown tasks~\cite{doke2021survey}. Various learning algorithms, such as support vector machines, linear regression, and neural networks, can be used under the meta-learning framework. SmartML, presented in~\cite{maher2019smart}, leverages a meta-learning framework to gather the knowledge and experience for automated selection and hyper-parameter tuning of machine learning algorithms. Specifically, in each task run, meta-knowledge is continuously added to improve the performance of the meta-learning framework further~\cite{maher2019smart}. In~\cite{lars2017auto}, Auto-Weka based on Bayesian optimization is proposed for automatic model selection and hyper-parameter tuning. Auto-Sklearn, introduced in~\cite{feurer2015efficient}, is a framework that utilizes past performance on similar tasks to automatically select proper machine learning models.  


Autoencoder is a type of artificial neural network, and it comprises an encoder and a decoder~\cite{provotar2019unsupervised}. Autoencoder-based models can learn correlations between features of normal data and detect data that are not from the normal distribution as anomalies. Inspired by~\cite{akcay2019ganomaly, tol2017adagan, Han_Chen_Liu_2021, provotar2019unsupervised}, we propose an encoder-decoder-encoder based ensemble model for anomaly detection on power grids to reduce the impact caused by the hacking problem. Like GANs~\cite{akcay2019ganomaly}, only normal data are used as the input in the training phase. Our proposed model also has a generator constructed by the first encoder and decoder, whereas, unlike GANs, our model does not include a discriminator. In our approach, the first encoder maps inputs into latent representations, and the decoder reconstructs samples from the latent space to the original space. We employ another encoder to map reconstructed samples into the latent space to learn the distribution of latent representations. The structures of the two encoders in our approach are the same. In our proposed ensemble learning framework, multiple encoder-decoder-encoder networks and sample re-weighting are leveraged to model normal distribution better. Moreover, we explore the effectiveness of employing long-short term memory (LSTM) as encoders and decoders and investigate a meta-learning framework for hyper-parameter selection. Hence, our proposed ensemble method can detect anomalies more accurately and promptly, minimizing the harmful effects of hacking and fake data injections.

The main contributions of this work are as follows:
\begin{itemize}
\item we introduce a novel encoder-decoder-encoder-based ensemble framework with sample re-weighting for anomaly detection on power grids.
\item we investigate a meta-learning framework for automatic hyper-parameter tuning.
\item The effectiveness of the proposed framework has been showcased on two real-world data sets.
\end{itemize}

The rest of this paper is organized as follows. Section II presents the technical background. The proposed encoder-decoder-encoder-based ensemble framework and meta-learning algorithm is discussed in Section III. Section IV presents experiments and results on the power system dataset and network intrusion dataset. Section V concludes this work.

%
%
%
%



 




\section{Technical Background}

\subsection{Anomaly Detection}
Anomaly detection has been extensively explored in many domains, such as computer vision tasks, disease marker discoveries, and computer security problems, as shown in~\cite{akcay2019ganomaly, zenati2018, schlegl2017unsupervised, AHMED201619}. Due to the lack of a large number of anomalies in the real world, viewing anomaly detection as a supervised learning task is not always feasible.
Anomaly detection can be formulated as a semi-supervised problem~\cite{Han_Chen_Liu_2021}, in which it is assumed that we have a set of normal samples $X = \{x_{n} \in \mathbb{R}^d: n = 1, ..., N \}$ as the training set and a test sample $\tilde{x} \in \mathbb{R}^d$ which may or may not come from the distribution of normal samples.
Anomaly detection aims to train a model to distinguish whether the sample $\tilde{x}$ is from the distribution of normal data. If $\tilde{x}$ is not from the normal data distribution, the model should classify it as an anomalous sample. Most models output an anomaly score for the input and label it anomalous if the score is higher than the pre-defined threshold~\cite{Han_Chen_Liu_2021}. Clustering methods, one-class classification models, nearest neighbor methods, and reconstructed-based approaches are employed to solve anomaly detection problems~\cite{akcay2019ganomaly, xiong2011group, chen2001one, zimek2012survey}. Recently, numerous works utilize deep neural networks, such as autoencoders and GANs, to solve anomaly detection tasks and make significant improvements in detecting anomalies~\cite{zhou2017anomaly, zong2018deep, schlegl2017unsupervised}. 

\subsection{Autoencoder} 
An autoencoder first compresses input data into a latent space and then generates reconstructed samples using latent vectors~\cite{Shen_Yu_Ma_Kwok_2021}. Autoencoders have been widely utilized in natural language process, time series prediction, and anomaly detection problems~\cite{Shen_Yu_Ma_Kwok_2021}. Specifically, an autoencoder comprises an encoder $E(\cdot; \phi)$ and a decoder $D(\cdot; \psi)$. Given an input sample $x$, the generated output of the autoencoder is $x' = D(E(x; \phi); \psi)$. The goal of training an autoencoder is to capture the nonlinear correlations among input features~\cite{chen2018auto}. For anomaly detection tasks, autoencoders model the data distribution of normal samples by minimizing the distance between the normal sample $x$ and the reconstructed sample $x'$. As shown in~\cite{zhou2017anomaly}, an autoencoder can solve the following optimization problem~\cite{zhou2017anomaly}: 
\begin{equation}\label{eq:1}
min_{D, E} ||x - D(E(x; \phi); \psi)||
\end{equation}
where $\phi$ and $\psi$ are the parameters for the encoder and decoder, respectively, and $l_2$-norm is usually utilized for $||\cdot||$. Compared with supervised approaches for anomaly detection~\cite{sakhnini2019}, employing autoencoder-based models for anomaly detection problems is efficient as only normal data are necessary for the training phase, and collecting various labeled anomalous data is costly and not always possible.

\subsection{Ensemble Autoencoder}

A single model may not have a stable and robust performance due to the high variance or bias. Ensemble learning combines several single-base learners to improve the overall performance of the model~\cite{zhu2020geometric}.  
In~\cite{han2020research, chen2018evo, Shen_Yu_Ma_Kwok_2021}, multiple autoencoders are employed for anomaly detection tasks. The autoencoder-based ensemble model can capture most correlations between input features and model the distribution of normal data, even for high-dimensional inputs~\cite{Shen_Yu_Ma_Kwok_2021}. Chen~\cite{chen2018evo} proposes an approach that combines the autoencoder with Adaboost (ADAE) for image anomaly detection. ADAE can capture more feature correlations than a single autoencoder by enabling successive autoencoders to focus on correlations the previous model failed to capture. Hence, an ensemble of autoencoders can better model the normal data distribution and detect anomalies more efficiently.

\section{Proposed Method}

This section details our proposed encoder-decoder-encoder networks-based ensemble anomaly detection framework and the meta-learning algorithm for automatic hyper-parameter tuning.
 
\subsection{Encoder-Decoder-Encoder Network for Anomaly Detection}

\begin{figure}[t]
    \centering
    \includegraphics[width=9.2cm]{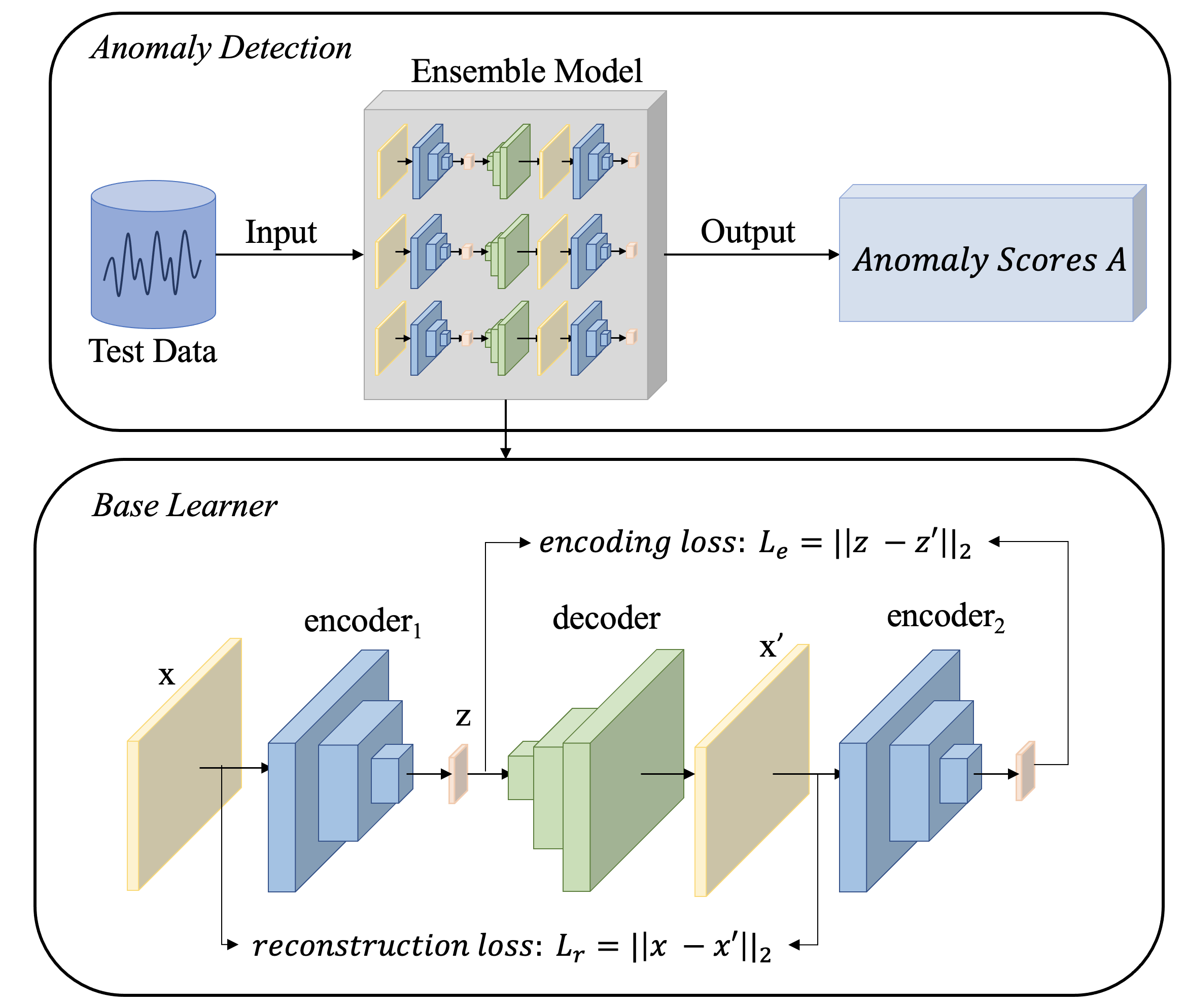}
    \caption{The ensemble framework of EDE network. The first part includes the overall framework of the ensemble method, and the second part describes the details of the EDE base learner. In the EDE base learner, $x$ is the input, and $x'$ is the reconstructed sample of $x$. $z$ and $z'$ are latent vectors of input $x$ and reconstructed sample $x'$, respectively.}
    \label{fig:figure1}
\end{figure}

The overview of our proposed encoder-decoder-encoder (EDE) base learner is illustrated in Figure~\ref{fig:figure1}. Since collecting anomalous data in the real world may be impracticable, only normal data are utilized as inputs in the training phase. The goal of the EDE model is to capture features of normal data as much as possible during the training phase. In the testing phase, the unique characteristics of the anomaly cannot pass through the EDE network, resulting in a large difference between the reconstructed sample and the anomaly. Then, our model assigns a large anomaly score to the anomalous sample. In EDE, the first encoder $E_1(\cdot;\phi)$ converts the input data $x \in \mathbb{R}^d $ into a latent representation $z \in \mathbb{R}^c $ which can well capture the features of $x$~\cite{akcay2019ganomaly}. The decoder $D(\cdot;\psi)$ then maps the latent vector $z$ back to the original dimension $\mathbb{R}^d$ and generates the reconstructed sample $x'$. The additional encoder $E_2(\cdot;\Tilde{\phi})$ has the same structure with $E_1(\cdot;\phi)$ but different parameters. $E_2(\cdot;\Tilde{\phi})$ encodes reconstructed sample $x'$ into the latent space and outputs its latent vector $z' \in \mathbb{R}^c$. Similar to~\cite{akcay2019ganomaly, Han_Chen_Liu_2021}, for training our EDE network, the reconstruction loss $L_r$ and encoding loss $L_e$ are minimized to capture features of normal data and latent representations. The reconstruction loss is: 
\begin{equation}\label{eq:2}
L_r(x) = ||x - D(E_1(x; \phi); \psi)||_2
\end{equation}
which computes the dissimilarity between real input $x$ and reconstructed sample $x'$. 
The encoding loss measures the difference between latent representations of the real input $x$ and reconstructed sample $x'$. The encoding loss is computed as:
\begin{equation}\label{eq:3}
L_e(x) = ||E_1(x; \phi) - E_2(x'; \Tilde{\phi})||_2
\end{equation}
The training objective of our proposed model is to minimize the combination of reconstruction loss and encoding loss:
\begin{equation}\label{eq:4}
min_{\phi, \psi, \Tilde{\phi}}\sum^N_{j=1}\alpha L_r(x_j;\phi, \psi) + \beta L_e(x_j;\phi, \Tilde{\phi})
\end{equation}

Based on this training process, unique features of anomalies cannot pass through the encoder-decoder part as EDE is only trained on normal data and can only capture normal features. For an anomaly sample $\Bar{x}$, its reconstruction $\Bar{x}'$ is very different from real $\Bar{x}$ resulting in a large difference between latent vectors $\Bar{z}$ and $\Bar{z}'$. Therefore, encoding loss $L_e$ can be used for detecting anomalies. For a test sample $\Tilde{x}$, the anomaly score $a(\Tilde{x})$ is:

\begin{equation}\label{eq:5}
a(\Tilde{x}) = ||E_1(\Tilde{x}) - E_2(D(E_1(\Tilde{x})))||_2
\end{equation}

Anomaly scores $A = \{a(\Tilde{x}_j) : \Tilde{x}_j \in D_{test}\}$ for test data $D_{test}$ are normalized into the range [0, 1] to evaluate the performance of our proposed model:

\begin{equation}\label{eq:6}
a'(\Tilde{x}_j) = \frac{a(\Tilde{x}_j) - min(A)}{max(A) - min(A)}
\end{equation}

\begin{algorithm}[t]\small
  \caption{Ensemble learning based on EDE networks}
    \label{alg:ede}
\textbf{Input:} Training dataset of normal samples $D_{train}$ ($D_{train} = \{x_1, x_2, ..., x_N\}$) with a uniform weight $W_0 = (1/N,...,1/N)$.
\begin{algorithmic}[1]
    \STATE Initialize parameters $\{(\phi_i, \psi_i, \Tilde{\phi}_i): i = 1, ..., I\}$ \\
    \FOR{$e =  1, ... , E$}
        \STATE $W_e = UpdateTrainingSampleWeights(EDE_{en}, D_{train})$  \\
        \FOR{$t = 1, ..., T$}
            \STATE Sample $i$ from $\{1, ...,i\}$ \\
            \STATE Sample a minibatch of training data $D^t_{train}$ from $D_{train}$ \\
            \text Train $(E_1(\cdot;\phi_i), D(\cdot;\psi_i), E_2(\cdot;\Tilde{\phi}_i))$ on $D^t_{train}$ with weight $W_e^t$ \\
            \STATE Calculate the reconstruction loss $L^i_r$ and encoding loss $L^i_e$. \\
            \STATE $L^i = \alpha L^i_r + \beta L^i_e$ \\
            \text Update parameters of $(E_1(\cdot;\phi_i), D(\cdot;\psi_i), E_2(\cdot;\Tilde{\phi}_i))$ \\
            \STATE Update $E_1(\cdot;\phi_i) : \phi_i \leftarrow \phi_i - \nabla_{\phi_i}L^i $ \\
            \STATE Update $D(\cdot;\psi_i) : \psi_i \leftarrow \psi_i - \nabla_{\psi_i}L^i $   \\
            \STATE Update $E_2(\cdot;\Tilde{\phi}_i) : \Tilde{\phi}_i \leftarrow \Tilde{\phi}_i - \nabla_{\Tilde{\phi}_i}L^i $  \\
        \ENDFOR
    \ENDFOR
  \end{algorithmic}
  \raggedright\textbf{Output:} Trained EDE based ensemble model $EDE_{en} = \{(E_1(\cdot;\phi_i), D(\cdot;\psi_i), E_2(\cdot;\Tilde{\phi}_i)): i = 1, ..., I\}$ 
\end{algorithm}

\subsection{Ensemble learning framework based on EDE}

In this work, we utilize the ensemble of our proposed model EDE for solving anomaly detection problems, and the framework is shown in Figure~\ref{fig:figure1}. Algorithm~\ref{alg:ede} presents the training phase of our proposed ensemble method. Multiple EDE networks $EDE_{en} = \{(E_1(\cdot;\phi_i), D(\cdot;\psi_i), E_2(\cdot;\Tilde{\phi}_i)): i = 1, ..., I\}$ are initialized with different parameters. Denoting the reconstruction loss and encoding loss for a single EDE network as $L_r^i(x) = L_r(x;\phi_i, \psi_i)$ and $L_e^i(x) = L_e(x;\phi_i, \Tilde{\phi}_i)$, our model minimizes the sum of all losses during the training phase:

\begin{equation}\label{eq:7}
min_{\{\phi, \psi, \Tilde{\phi}\}_{i=1}^I}\sum^I_{i=1}\alpha L_r^i + \beta L_e^i
\end{equation}
where $\alpha$ and $\beta$ are the weight of the reconstruction loss and the weight of encoding loss, respectively.
For each epoch, as the function in line 3 
in Algorithm~\ref{alg:ede}, weights of training data are updated based on the performance of the current ensemble model EDE$_{en}$. Specifically, since only normal data are utilized for the training phase, anomaly scores for training samples should be low, and samples with larger anomaly scores should be assigned with a larger weight. With re-weighting samples, the ensemble model can capture data modes and correlations between features of normal data. 
For each iteration within one epoch, only one of the base models is chosen to be updated. Specifically, a random EDE network is trained and updated in each iteration based on a random minibatch of training data. Similar to \cite{Han_Chen_Liu_2021}, this training process is particularly efficient because one base model, on average, is only trained once in $I$ iterations. Details of the training algorithm are in Algorithm~\ref{alg:ede}. The anomaly score of the ensemble model EDE$_{en}$ for a test sample $\Tilde{x}$ is the average of the scores of all base networks:

\begin{equation}\label{eq:8}
a(\Tilde{x}) = \frac{1}{I}\sum_{i=1}^IA(\Tilde{x}; \phi_i,\psi_i, \Tilde{\phi}_i)
\end{equation}

\begin{figure}[t]
    \centering
    \includegraphics[width=9.2cm]{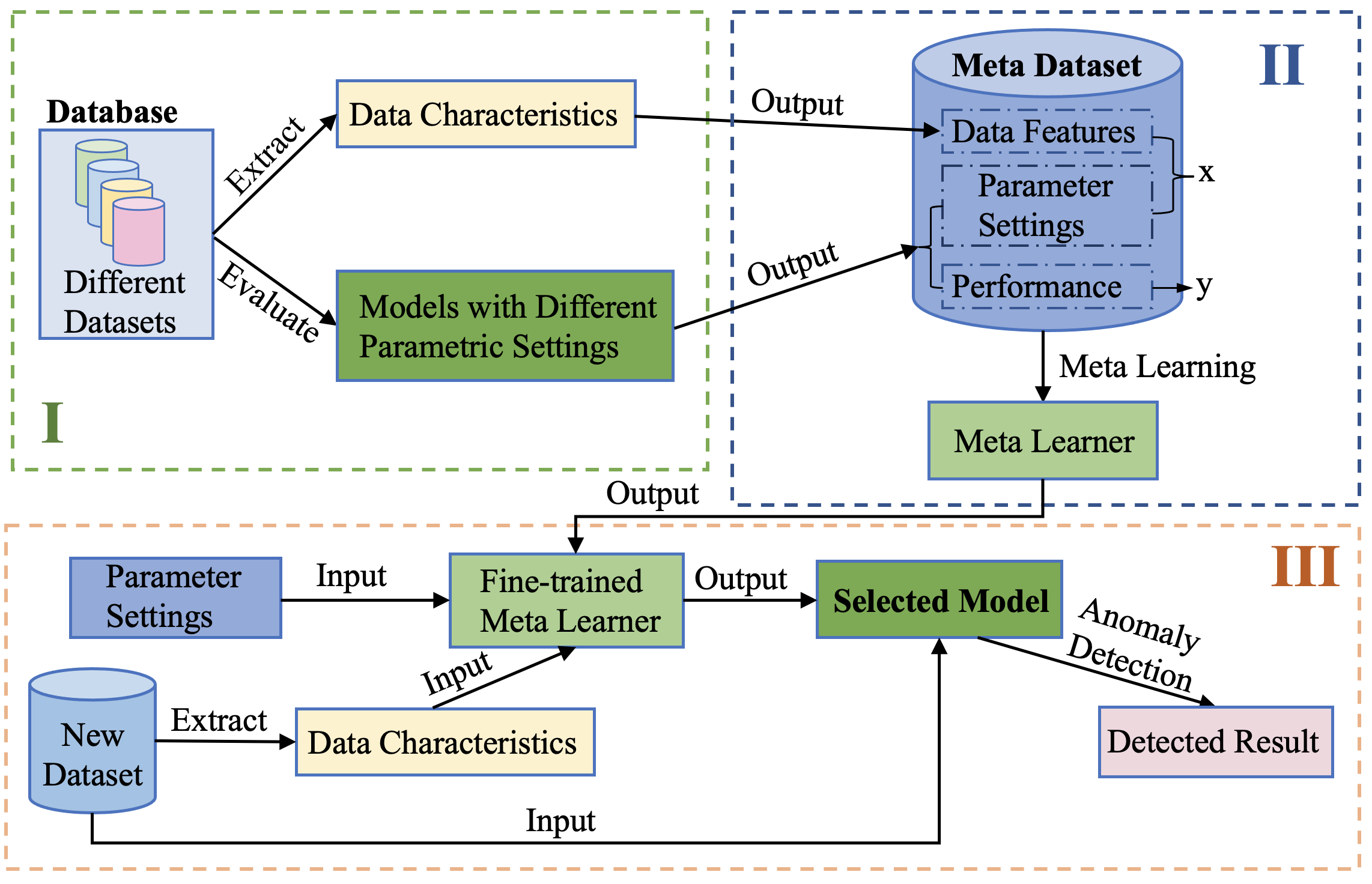}
    \caption{The overview of our proposed meta-learning algorithm. Phase I generates the meta data, and Phase II builds the meta dataset and trains the meta-learner. Finally, the meta-learner is tested on different new datasets in Phase III.}
    \label{fig:figure2}
\end{figure}

\subsection{Meta-learning algorithm for automatic hyper-parameter tuning}
In order to choose an ensemble model with the proper parametric setting, we employ a meta-learning algorithm to tune the hyper-parameters automatically. The overall structure of the meta-learning algorithm is illustrated in Figure~\ref{fig:figure2}. In Phase I, various datasets and models with different parameter-settings are used to generate meta data. Specifically, data features, including the number of instances, the number of sparse features, the number of positive skewness features, and the number of negative skewness features, are extracted from various tasks. A feature is considered sparse if zeros account for more than half of all values of the feature. The skewness of features is calculated using Pearson mode with the following formula:

\begin{equation}\label{eq:9}
Skewness = \frac{3(mean - median)}{\sigma}
\end{equation}
where mean, median, and $\sigma$ are the average, median, and standard deviation of the values of an input feature, respectively. We train and evaluate each model on different datasets and obtain their performance. In Phase II, meta data is collected from the outputs of Phase I. Meta dataset consists of data features of different datasets, various parametric settings of models, and the performance of models on different datasets. Then, the meta-learner is trained with the generated meta data to get some prior knowledge. When testing on a new task in Phase III, the meta-learner utilizes information obtained before and the extracted data features from the current task to predict the performance of each model on this dataset. Finally, the meta-learning framework advises which model should be used on the current task. The proposed meta-learning framework does not have constraints on the type of meta-learners, and it can be used to determine the values of all kinds of hyper-parameters. We choose support vector regression (SVR) as the meta-learner in our approach, and we employ this framework to automatically determine the number of EDE-based models.

\section{Experiments}
This section first introduces the datasets used for anomaly detection tasks, different baselines, and metrics for evaluating all models. Then, we test and compare the performance of our models and other baseline methods. Finally, we investigate the effectiveness of utilizing LSTM as encoders and decoders and the automatic hyper-parameter tuning of the meta-learning framework. 

\begin{table*}[htp] 
\begin{center}
\begin{tabular}{m{1.8cm} | m{2cm}| m{2cm} | m{2cm}| m{2cm} | m{2cm}}
 \hline
\textbf{Method} & \textbf{Precision} & \textbf{Recall} & \textbf{$F_1$ score} & \textbf{Accuracy} & \textbf{AUROC} \\ 
 \hline
\textbf{OC-SVM} & 0.745 $\pm$ 0.004 & 0.849 $\pm$ 0.005 & 0.792 $\pm$ 0.004 & 0.897 $\pm$ 0.003 & 0.823 $\pm$ 0.003\\
 \hline
\textbf{FNN} & 0.557 $\pm$ 0.123 & 0.511 $\pm$ 0.142 & 0.526 $\pm$ 0.129 & 0.810 $\pm$ 0.045 & 0.665 $\pm$ 0.150\\
 \hline
\textbf{DAGMM} & 0.923 $\pm$ 0.001 & 0.938 $\pm$ 0.001 & 0.930 $\pm$ 0.001 & 0.972 $\pm$ 0.001 & 0.989 $\pm$ 0.020\\
 \hline
\textbf{AdaGAN} & 0.836 $\pm$ 0.007 & 0.829 $\pm$ 0.004 & 0.833 $\pm$ 0.005 & 0.933 $\pm$ 0.003 & 0.864 $\pm$ 0.007\\
 \hline
\textbf{GANomaly} & 0.933 $\pm$ 0.010 & 0.993 $\pm$ 0.024 & 0.962 $\pm$ 0.015 & 0.984 $\pm$ 0.035 & 0.992 $\pm$ 0.173\\
 \hline
\textbf{EDE} & 0.956 $\pm$ 0.011 & 0.972 $\pm$ 0.013 & 0.964 $\pm$ 0.012 & 0.986 $\pm$ 0.015 & 0.995 $\pm$ 0.170\\
 \hline
\textbf{GANomaly$_{en}$} & 0.935 $\pm$ 0.009 & \textbf{0.995 $\pm$ 0.011} & 0.964 $\pm$ 0.010 & 0.985 $\pm$ 0.009 & 0.996 $\pm$ 0.095\\
 \hline
\textbf{EDE$_{en}$} & \textbf{0.966 $\pm$ 0.013} & 0.983 $\pm$ 0.037 & \textbf{0.974 $\pm$ 0.022} & \textbf{0.990 $\pm$ 0.015} & \textbf{0.998 $\pm$ 0.191}\\
 \hline

\end{tabular}
\caption{Experimental results on the KDD99 dataset. Performance comparison of our approaches and other baselines. Our ensemble model EDE$_{en}$ outperforms baselines.}
\label{table:1}
\end{center}
\end{table*}

\begin{table*}[htp] 
\begin{center}
\begin{tabular}{m{1.8cm} | m{2cm}| m{2cm} | m{2cm}| m{2cm} | m{2cm}}
 \hline
\textbf{Method} & \textbf{Precision} & \textbf{Recall} & \textbf{$F_1$ score} & \textbf{Accuracy} & \textbf{AUROC} \\ 
 \hline
\textbf{OC-SVM} & 0.361 $\pm$ 0.005 & 0.495 $\pm$ 0.007 & 0.418 $\pm$ 0.005 & 0.497 $\pm$ 0.004 & 0.496 $\pm$ 0.004\\
 \hline
\textbf{FNN} & 0.276 $\pm$ 0.137 & 0.046 $\pm$ 0.035 & 0.070 $\pm$ 0.043 & 0.770 $\pm$ 0.027 & 0.502 $\pm$ 0.028\\
 \hline
\textbf{DAGMM} & 0.232 $\pm$ 0.008 & 0.194 $\pm$ 0.007 & 0.212 $\pm$ 0.007 & 0.653 $\pm$ 0.004 & 0.505 $\pm$ 0.007\\
 \hline
\textbf{AdaGAN} & 0.280 $\pm$ 0.009 & 0.243 $\pm$ 0.004 & 0.260 $\pm$ 0.006 & 0.681 $\pm$ 0.008 & 0.528 $\pm$ 0.010\\
 \hline
\textbf{GANomaly} & 0.454 $\pm$ 0.026 & 0.362 $\pm$ 0.034 & 0.427 $\pm$ 0.029 & 0.796 $\pm$ 0.013 & 0.709 $\pm$ 0.035\\
 \hline
\textbf{EDE} & 0.470 $\pm$ 0.024 & 0.568 $\pm$ 0.047 & 0.513 $\pm$ 0.032 & 0.857 $\pm$ 0.017 & 0.784 $\pm$ 0.020\\
 \hline
\textbf{GANomaly$_{en}$} & 0.490 $\pm$ 0.018 & 0.389 $\pm$ 0.020 & 0.433 $\pm$ 0.019 & 0.808 $\pm$ 0.025 & 0.771 $\pm$ 0.034 \\
 \hline
\textbf{EDE$_{en}$} & \textbf{0.499 $\pm$ 0.017} & \textbf{0.576 $\pm$ 0.024} & \textbf{0.535 $\pm$ 0.021} & \textbf{0.858 $\pm$ 0.010} & \textbf{0.823 $\pm$ 0.017}\\
 \hline

\end{tabular}
\caption{Experimental results on the ICS Cyber Attack dataset. Comparison of precision, recall, $F_1$ score, accuracy, and AUROC scores for our approaches and other baselines. The performance of our ensemble model EDE$_{en}$ is the best.}
\label{table:2}
\end{center}
\end{table*}

\subsection{Datasets}
We evaluate our approaches and baselines on two datasets: KDD99 dataset\footnote{http://kdd.ics.uci.edu/databases/kddcup99/kddcup99.html} and a standard simulated industrial control system (ICS) cyber attack dataset\footnote{https://sites.google.com/a/uah.edu/tommy-morris-uah/ics-data-sets}. 

\subsubsection{KDD99} KDD99 dataset in the UCI repository~\cite{lichman2013UCI} is a network intrusion dataset containing 41 features for anomaly detection. Due to the large size of the original KDD99 dataset, we use KDD99 10 percent dataset in all experiments following the instructions in the UCI repository and the experimental setup in~\cite{zenati2018}. Similar to~\cite{zenati2018}, seven categorical features are converted to numerical features via the one-hot encoding method. After this process, each sample is a 121-dimensional vector. Because of the ratio of anomalous data to normal data, we follow the setup of~\cite{zenati2018} on KDDCUP99 10 percent dataset: treat normal data as anomalous data. In this way, we classify test samples with anomaly scores in the top 20\% as anomalies, and other data pre-process for this dataset is based on the experimental setup of~\cite{zenati2018} as well.

\subsubsection{ICS Cyber Attack Dataset} This power dataset~\cite{borge2014ml} contains 15 subsets, including 37 power system events which consist of 28 attack events and nine normal events. Multiple types of attack events, such as remote tripping command injection, relay setting change, and data injection, are included in this dataset. Each sample in this dataset has 128 features. Due to the proportion of anomalies and normal data in this power system dataset, we randomly select attack events accounting for 20\% of total events in each subset to test all approaches. A random 80\% of each sub-dataset is selected as the training set and the rest as the testing set. All experimental results on the power dataset are the average results on its subsets. In the experiment to explore the effectiveness of the meta-learning framework, ten subsets are randomly selected to build the meta database to train the meta-learner, and the remaining subsets are used as the testing set.

\subsection{Baselines and Metrics}

\subsubsection{Baselines}
Traditional machine learning methods, OC-SVM~\cite{chen2001one} and Feed-forward Neural Network (FNN)~\cite{bebis1994feed} are considered as basic baseline methods. Some recent works for anomaly detection, Deep Autoencoding Gaussian Mixture Model (DAGMM)~\cite{zong2018deep}, GANomaly~\cite{akcay2019ganomaly}, and GANomaly Ensemble~\cite{Han_Chen_Liu_2021}, are employed as baselines as well. Besides, we also adapt AdaGAN~\cite{tol2017adagan} for anomaly detection tasks. The details for baselines are described below:
\begin{itemize}
\item OC-SVM. This is a traditional unsupervised method used for anomaly detection problems. In this work, the radial basis function (RBF) kernel is employed in all experiments.

\item FNN. We employ a feed-forward neural network model including three hidden layers with 64, 64, and 32 as the number of hidden nodes for each layer, respectively. Tanh is used as the activation function, and Adam is used as the optimizer.

\item DAGMM. DAGMM consisting of a compression network and estimation network~\cite{zong2018deep} is a state-of-the-art machine learning method for unsupervised anomaly detection. The implementation of DAGMM is based on the work~\cite{zong2018deep}.

\item AdaGAN. This method combines the AdaBoost algorithm with GAN models to cover the distribution of input data~\cite{tol2017adagan}. We adapt this technique for anomaly detection by utilizing the pre-defined threshold to detect anomalies. Specifically, samples with scores higher than 80 percent of all output scores would be labeled anomalies. The implementation of AdaGAN for this experiment is based on~\cite{tol2017adagan}.

\item GANomaly. This is an alternative GAN model employing encoder-decoder-encoder sub-networks as the generator, and the implementation of the GANomaly model for all tasks is the same with~\cite{akcay2019ganomaly}.

\item GANomaly Ensemble. This is a GANomaly-based ensemble model in which a group of generators and discriminators are jointly trained~\cite{Han_Chen_Liu_2021}. Specifically, each generator gets feedback from all discriminators, and each discriminator collects reconstructed samples from all generators. We implement the GANomaly Ensemble model based on the work~\cite{Han_Chen_Liu_2021}.

\end{itemize}

\subsection{Experimental Setup and Comparision Metrics}
For our proposed models EDE and EDE$_{en}$, encoders and decoders are implemented with feed-forward neural networks including three fully connected layers and hyperbolic tangent function (Tanh) as the activation function, respectively. LSTM-EDE$_{en}$ and MetaLSTM-EDE$_{en}$ models on the KDD99 dataset use an LSTM with one recurrent layer, while both models on the ICS cyberattack dataset employ the LSTM with two recurrent layers. For EDE$_{en}$ and LSTM-EDE$_{en}$, $I = 3$ is the number of EDE-based models. In addition, we also set 3 as the number of base models for the baseline GANomaly Ensemble. In this manner, we can compare our proposed ensemble model and this baseline fairly. For the MetaLSTM-EDE$_{en}$ model, our proposed meta-learning framework determines the number of its EDE-based models.

We consider precision, recall, $F_1$ score, and accuracy as metrics to evaluate our approaches and baseline methods. Additionally, the area under Receiver Operating Characteristic (AUROC) curve is also employed as one metric since it avoids defining the threshold for anomaly scores of test samples when evaluating models.

\begin{figure}[t]
    \centering
    \includegraphics[width=8cm]{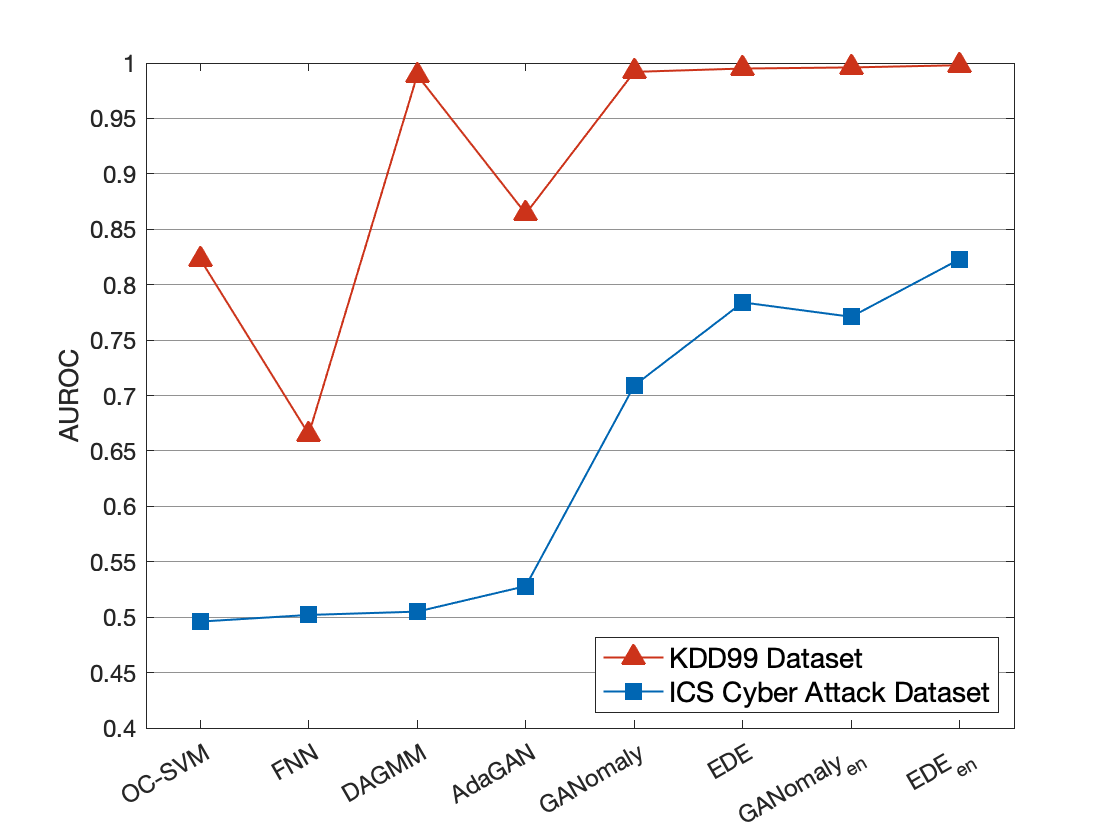}
    \caption{AUROC scores for all approaches on KDD99 dataset and ICS cyber attack dataset. EDE$_{en}$ achieves the highest AUROC score.}
    \label{fig:figure3}
\end{figure}

\subsection{Experimental Results and Discussion}

\begin{figure}
     \centering
     \begin{subfigure}[b]{0.21\textwidth}
         \includegraphics[width=\textwidth]{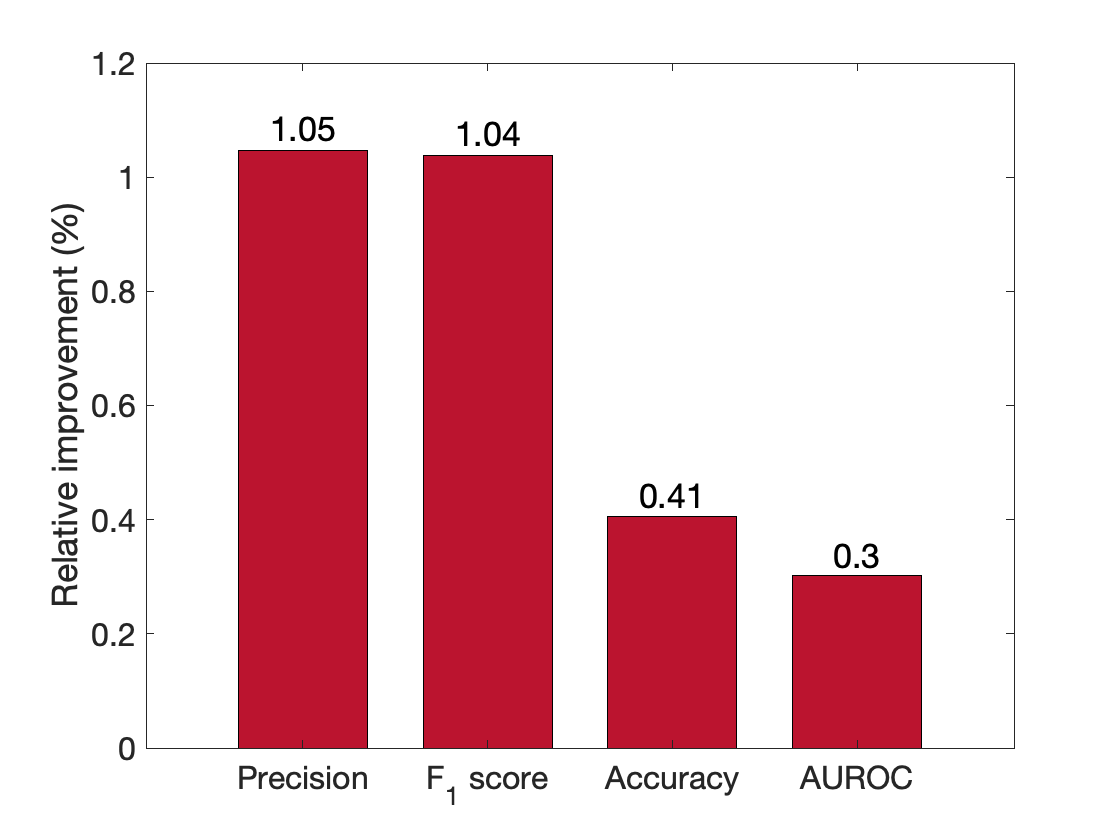}
         \caption{EDE on the KDD99 dataset}
         \label{fig:kddede}
     \end{subfigure}
     \begin{subfigure}[b]{0.21\textwidth}
         \includegraphics[width=\textwidth]{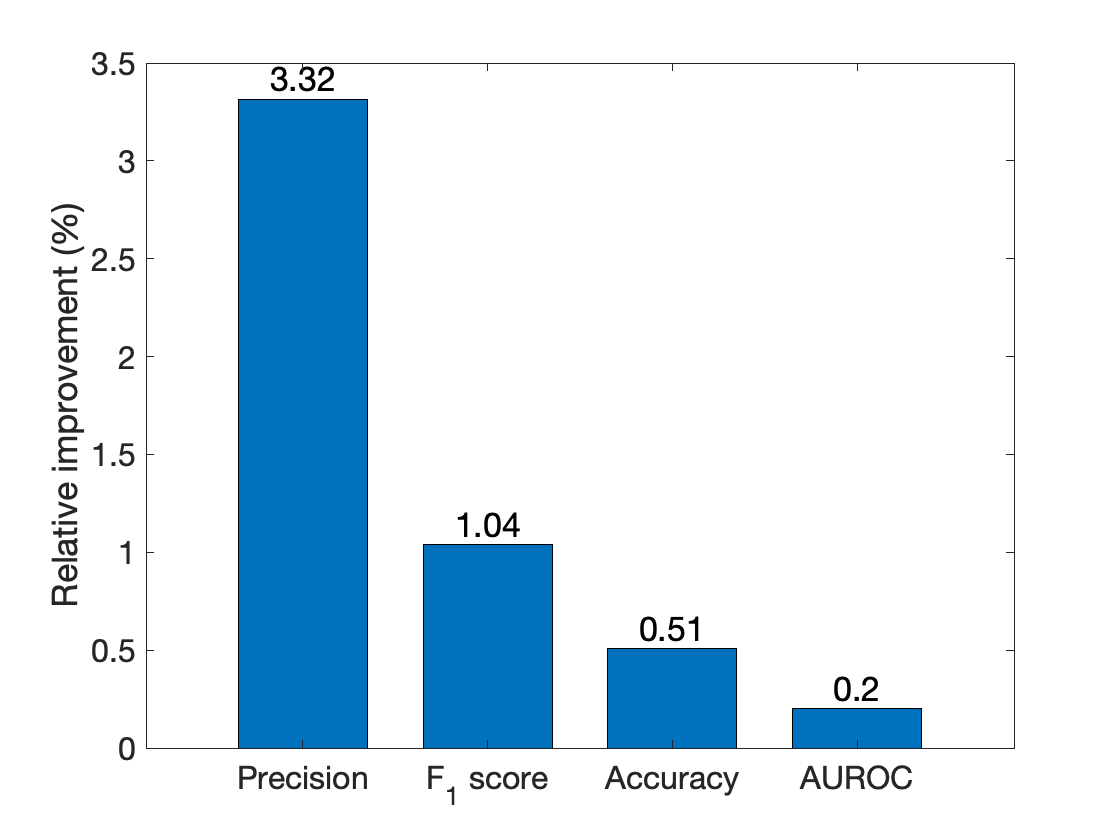}
         \caption{GANomaly$_{en}$ on the KDD99 dataset}
         \label{fig:kddgan}
     \end{subfigure}
     \hspace{0mm}
     \begin{subfigure}[b]{0.21\textwidth}
         \includegraphics[width=\textwidth]{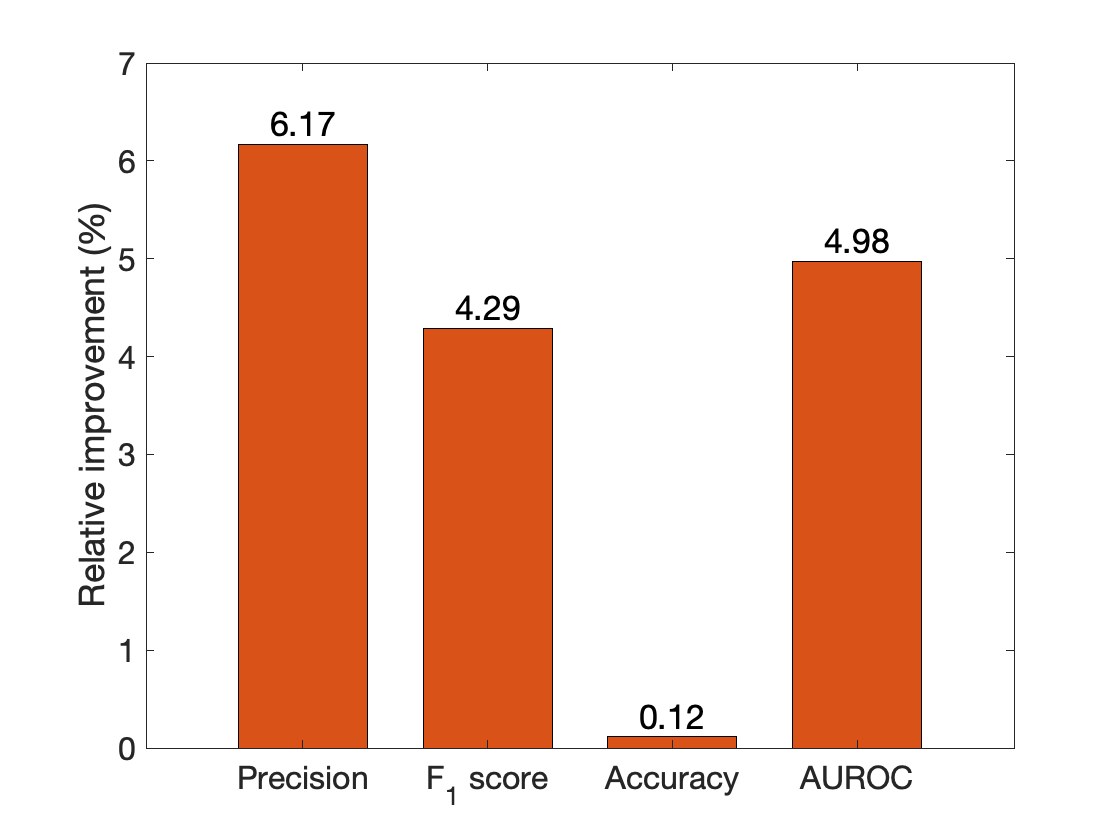}
         \caption{EDE on the ICS cyber attack dataset}
         \label{fig:icsede}
     \end{subfigure}
     \begin{subfigure}[b]{0.21\textwidth}
         \includegraphics[width=\textwidth]{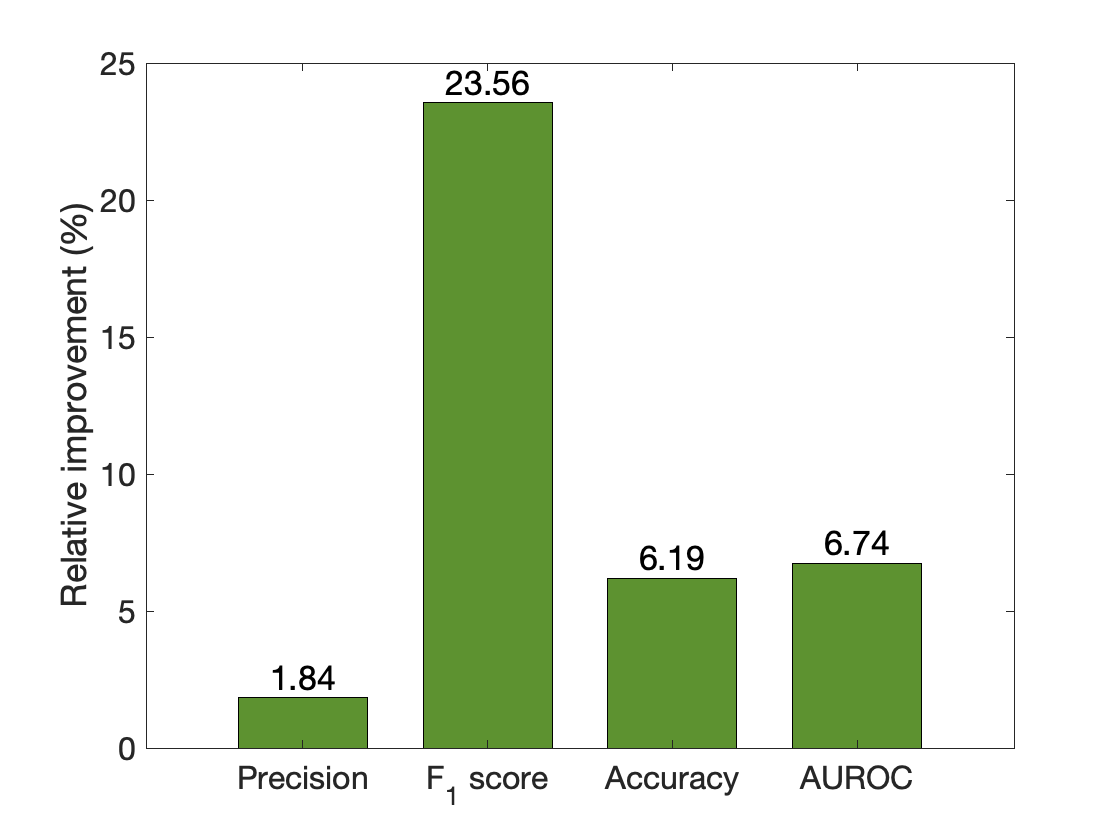}
         \caption{GANomaly$_{en}$ on the ICS cyber attack dataset}
         \label{fig:icsgan}
     \end{subfigure}
     
        \caption{Relative improvement of our ensemble model EDE$_{en}$ over its single model EDE and the best baseline GANomaly$_{en}$ on both KDD99 dataset and ICS cyber attack dataset.}
        \label{fig:figure4}
\end{figure}

\begin{figure}[t]
    \centering
    \includegraphics[width=9cm]{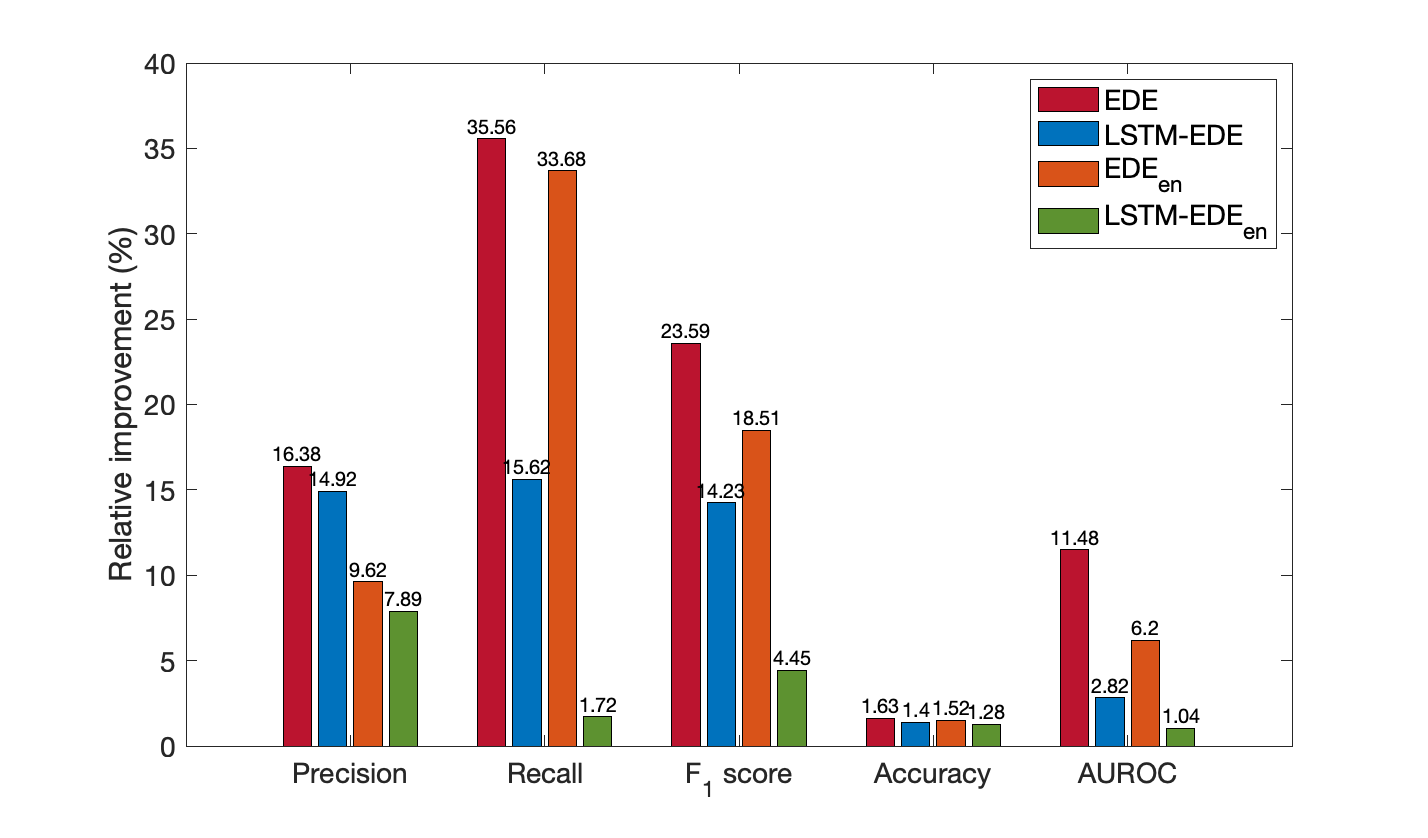}
    \caption{Relative improvement of MetaLSTM-EDE$_{en}$ over our other proposed approaches on ICS cyber attack dataset.}
    \label{fig:figure5}
\end{figure}

In this subsection, we test and discuss the performance of our various approaches and all baselines on the KDD99 dataset and ICS Cyber Attack Dataset. 
The highest scores for all metrics in Table~\ref{table:1}, Table~\ref{table:2}, and Table~\ref{table:3} are marked in bold. In the first two experiments, our proposed ensemble EDE$_{en}$ outperforms all baselines. Figure~\ref{fig:figure4} shows the relative improvement of EDE$_{en}$ over its single model EDE and the baseline GANomaly$_{en}$ on two datasets, respectively. From Figure~\ref{fig:figure4}, it is shown that our proposed model EDE$_{en}$ can outperform its single model EDE and the best baseline GANomaly$_{en}$ by all metrics except recall. Specifically, EDE$_{en}$ outperforms its single model EDE by 1.04\% in $F_1$ score on the KDD99 dataset and outperforms EDE by 6.17\% in precision on the ICS cyber attack dataset. In addition, EDE$_{en}$ outperforms the best baseline GANomaly$_{en}$ on precision by 3.32\% on the KDD99 dataset and outperforms GANomaly$_{en}$ on precision by 1.84\% and $F_1$ score by 23.56\% on the power dataset. Furthermore, in the last experiment, we compare the performance of the proposed models using feed-forward neural networks and models with LSTM as encoders and decoders. The effectiveness of using the meta-learning framework for determining hyper-parameters MetaLSTM-EDE$_{en}$ is also explored on the ICS cyber attack dataset. Figure~\ref{fig:figure5} illustrates the relative improvement of MetaLSTM-EDE$_{en}$ on our other proposed approaches. The results show that MetaLSTM-EDE$_{en}$ outperforms other approaches on all metrics, illustrating the meta-learning framework's robustness for automatic hyper-parameter tuning.

\subsubsection{Performance Evaluation of Models on the KDD99 Dataset}
Table~\ref{table:1} presents and compares the performance of our approaches and other baseline models, and Figure~\ref{fig:figure3} shows AUROC scores of all approaches on the KDD99 dataset. We observe that our proposed model EDE$_{en}$ outperforms the single model EDE and other baselines by obtaining the best scores of all metrics except the metric recall. Moreover, Table~\ref{table:1} demonstrates that the baseline GANomaly outperforms baselines OC-SVM, FNN, and DAGMM by a large margin on all metrics. This shows that using the reconstruction loss and encoding loss to optimize parameters gives the model a more robust capability of capturing correlations among input features. AdaGAN performs worse than DAGMM and GANomaly$_{en}$ since it is initially not designed for anomaly detection problems. The baseline GANomaly$_{en}$ performs better than a single GANomaly model, and our ensemble approach EDE$_{en}$ outperforms the single model EDE, verifying that the ensemble algorithm is very efficient and necessary for anomaly detection tasks. The EDE model is particularly robust since it reaches a similar $F_1$ score, accuracy score, and AUROC score with the ensemble baseline GANomaly$_{en}$. It illustrates that re-weighting normal samples for each epoch during the training phase is helpful as the model can focus on missed key correlations between input features. 

\begin{table*}[htp] 
\begin{center}
\begin{tabular}{m{2.5cm} | m{2cm}| m{2cm} | m{2cm}| m{2cm} | m{2cm}}
 \hline
\textbf{Method} & \textbf{Precision} & \textbf{Recall} & \textbf{$F_1$ score} & \textbf{Accuracy} & \textbf{AUROC} \\ 
 \hline
\textbf{EDE} & 0.470 $\pm$ 0.024 & 0.568 $\pm$ 0.047 & 0.513 $\pm$ 0.032 & 0.857 $\pm$ 0.017 & 0.784 $\pm$ 0.020\\
 \hline
\textbf{LSTM-EDE} & 0.476 $\pm$ 0.032 & 0.666 $\pm$ 0.025 & 0.555 $\pm$ 0.019 & 0.859 $\pm$ 0.010 & 0.850 $\pm$ 0.012\\
 \hline
\textbf{EDE$_{en}$} & 0.499 $\pm$ 0.017 & 0.576 $\pm$ 0.024 & 0.535 $\pm$ 0.021 & 0.858 $\pm$ 0.010 & 0.823 $\pm$ 0.017\\
 \hline
\textbf{LSTM-EDE$_{en}$} & 0.507 $\pm$ 0.023 & 0.757 $\pm$ 0.035 & 0.607 $\pm$ 0.028 & 0.860 $\pm$ 0.009 & 0.865 $\pm$ 0.011\\
 \hline
 \textbf{MetaLSTM-EDE$_{en}$} & \textbf{0.547 $\pm$ 0.017} & \textbf{0.770 $\pm$ 0.020} & \textbf{0.634 $\pm$ 0.019} & \textbf{0.871 $\pm$ 0.025} & \textbf{0.874 $\pm$ 0.034}\\
 \hline

\end{tabular}
\caption{Experimental results of our approaches on the ICS Cyber Attack dataset. Comparison of precision, recall, $F_1$ score, accuracy, and AUROC scores for our different methods. The performance of the proposed model MetaLSTM-EDE$_{en}$ is the best.}
\label{table:3}
\end{center}
\end{table*}

\subsubsection{Performance Evaluation of Models on the ICS Cyber Attack Dataset}
The performance of all methods on the ICS Cyber Attack Dataset is reported in Table~\ref{table:2} and Figure~\ref{fig:figure3}. Overall, models have worse performance on this dataset than on the KDD99 dataset, which indicates that detecting anomalies on the ICS Cyber Attack Dataset is more complicated. This is because the power dataset is much smaller than the KDD99 dataset and includes various attacks. The standard deviation of the AUROC score for all methods on this dataset is smaller than that on the KDD99 dataset, which the small size of the power dataset may cause. From Table~\ref{table:2}, we detect that the performance of DAGMM and AdaGAN is worse than that of OC-SVM. This implies that these two baselines are not capable of capturing correlations among features of input data from this relatively small power dataset. Moreover, it is manifested that our ensemble method EDE$_{en}$ demonstrates superior performance over other baselines concerning all metrics on the power dataset. EDE$_{en}$ outperforms the baseline GANomaly$_{en}$ on recall by 48.07\% and on AUROC by 6.74\%, which demonstrates that EDE$_{en}$ is more efficient on a smaller and more difficult dataset. Meanwhile, EDE also achieves a very high recall score, and it has the second-best performance in terms of all metrics except precision, further indicating the efficiency of re-weighting samples during the training phase. The ensemble methods GANomaly$_{en}$ and EDE$_{en}$ still achieve better performance than their single base models GANomaly and EDE showing the necessity and robustness of ensemble algorithms for detecting various cyber attacks.

\subsubsection{Investigation of the Effectiveness of the LSTM-based $EDE_{en}$ model and the meta-learning approach}
In this experiment, we first investigate the performance of EDE-based models with the LSTM as encoders and decoders on the ICS Cyber Attack Dataset. Then, the effectiveness of utilizing a meta-learning framework to determine hyper-parameters is explored. Specifically, the proposed meta-learning framework is employed to select the number of base learners in the ensemble model from \{1, 3, 5, 7, 10, 15\}. We test and gain the average performance of MetaLSTM-EDE$_{en}$ on the testing sets. The experimental results of MetaLSTM-EDE$_{en}$ and our other proposed approaches are presented in Table~\ref{table:3} and Figure~\ref{fig:figure5}. The single model LSTM-EDE demonstrates better performance than EDE, and the ensemble model LSTM-EDE$_{en}$ outperforms the model EDE$_{en}$ on all metrics, illustrating the effectiveness of employing LSTM to capture the correlations between input features. Moreover, it is manifested that MetaLSTM-EDE$_{en}$ outperforms all other models regarding all metrics. From Figure~\ref{fig:figure5}, we observe that MetaLSTM-EDE$_{en}$ outperforms the second best model LSTM-EDE$_{en}$ on $F_1$ score by 4.45\% and recall by 1.72\%, further showing the robustness of automatic hyper-parameter tuning using meta-learning framework instead of tuning by human experts.

\section{Conclusion}

Network intrusions and various cyber attacks on power systems have gained increasing attention in recent years. Anomaly detection is imperative and critical for maintaining stable operations on power grids. Hence, designing an efficient model for detecting anomalies on power grids is of significant importance. However, it is difficult to collect data for various types of cyber attacks in the real world, and new types of attacks are also emerging. This makes training traditional machine learning models particularly hard since there may not be enough anomalous samples for training models. This paper proposes a novel encoder-decoder-encoder model for detecting cyber attacks on smart grids and introduces ensemble learning and re-weighting samples to our approach. Furthermore, we introduce LSTM as a component of our approach and explore the effectiveness of employing the meta-learning framework to determine hyper-parameters. Extensive experiments have been implemented on two real-world datasets, and results show that our approach MetaLSTM-EDE$_{en}$ reaches the best performance over all baselines and our other approaches. The results illustrate the effectiveness of the ensemble framework in capturing the correlations between input features and the robustness of the meta-learning algorithm for automatic hyper-parameter tuning. In the real world, some anomalies may mimic normal data by changing a feature slightly, and our model may fail to detect this. In the future, we aim to further optimize our model to learn correlations between the input features to detect the aforementioned anomalies. We will also utilize the meta-learning framework to decide more hyper-parameters in our model. Meanwhile, we will implement our approach on more anomaly detection tasks, including time series data and image data, 
to investigate the robustness of our proposed method.

\bibliographystyle{IEEEtran}
\bibliography{ref}

\begin{thebibliography}{10}
\providecommand{\url}[1]{#1}
\csname url@samestyle\endcsname
\providecommand{\newblock}{\relax}
\providecommand{\bibinfo}[2]{#2}
\providecommand{\BIBentrySTDinterwordspacing}{\spaceskip=0pt\relax}
\providecommand{\BIBentryALTinterwordstretchfactor}{4}
\providecommand{\BIBentryALTinterwordspacing}{\spaceskip=\fontdimen2\font plus
\BIBentryALTinterwordstretchfactor\fontdimen3\font minus
  \fontdimen4\font\relax}
\providecommand{\BIBforeignlanguage}[2]{{%
\expandafter\ifx\csname l@#1\endcsname\relax
\typeout{** WARNING: IEEEtran.bst: No hyphenation pattern has been}%
\typeout{** loaded for the language `#1'. Using the pattern for}%
\typeout{** the default language instead.}%
\else
\language=\csname l@#1\endcsname
\fi
#2}}
\providecommand{\BIBdecl}{\relax}
\BIBdecl

\bibitem{yadav2016review}
S.~A. Yadav, S.~R. Kumar, S.~Sharma, and A.~Singh, ``A review of possibilities
  and solutions of cyber attacks in smart grids,'' in \emph{2016 International
  Conference on Innovation and Challenges in Cyber Security
  (ICICCS-INBUSH)}.\hskip 1em plus 0.5em minus 0.4em\relax IEEE, 2016, pp.
  60--63.

\bibitem{adiban2021step}
M.~Adiban, A.~Safari, and G.~Salvi, ``Step-gan: A one-class anomaly detection
  model with applications to power system security,'' in \emph{ICASSP 2021 -
  2021 IEEE International Conference on Acoustics, Speech and Signal Processing
  (ICASSP)}, 2021, pp. 2605--2609.

\bibitem{wu2017two}
D.~Wu, H.~Zeng, C.~Lu, and B.~Boulet, ``Two-stage energy management for office
  buildings with workplace ev charging and renewable energy,'' \emph{IEEE
  Transactions on Transportation Electrification}, vol.~3, no.~1, pp. 225--237,
  2017.

\bibitem{wu2018machine}
D.~Wu, \emph{Machine learning algorithms and applications for sustainable smart
  grid}.\hskip 1em plus 0.5em minus 0.4em\relax McGill University (Canada),
  2018.

\bibitem{ADIBAN2020101105}
M.~Adiban, H.~Sameti, and S.~Shehnepoor, ``Replay spoofing countermeasure using
  autoencoder and siamese networks on asvspoof 2019 challenge,'' \emph{Computer
  Speech \& Language}, vol.~64, p. 101105, 2020.

\bibitem{havlena2022accurate}
V.~Havlena, P.~Matoušek, O.~Ryšavý, and L.~Holík, ``Accurate automata-based
  detection of cyber threats in smart grid communication,'' \emph{IEEE
  Transactions on Smart Grid}, pp. 1--1, 2022.

\bibitem{tong2022fast}
N.~Tong, X.~Zeng, K.~Sun, Y.~Wang, and J.~Luo, ``A fast protection relay for
  mmc-based dc grids robust to cyber-physical anomalies,'' \emph{IEEE
  Transactions on Smart Grid}, pp. 1--1, 2022.

\bibitem{zhangde2021tecting}
Y.~Zhang, J.~Wang, and B.~Chen, ``Detecting false data injection attacks in
  smart grids: a semi-supervised deep learning approach,'' \emph{IEEE
  Transactions on Smart Grid}, vol.~12, no.~1, pp. 623--634, 2021.

\bibitem{mohammad2022anomaly}
A.~Mohammad~Saber, A.~Youssef, D.~Svetinovic, H.~H. Zeineldin, and E.~F.
  El-Saadany, ``Anomaly-based detection of cyberattacks on line current
  differential relays,'' \emph{IEEE Transactions on Smart Grid}, vol.~13,
  no.~6, pp. 4787--4800, 2022.

\bibitem{adiban2018}
M.~Adiban, H.~Sameti, N.~Maghsoodi, and S.~Shahsavari, ``Sut system description
  for anti-spoofing 2017 challenge,'' in \emph{Proceedings of the 29th
  Conference on Computational Linguistics and Speech Processing ({ROCLING}
  2017)}, 2017, pp. 264--275.

\bibitem{zhang2021time}
J.~E. Zhang, D.~Wu, and B.~Boulet, ``Time series anomaly detection for smart
  grids: A survey,'' in \emph{2021 IEEE Electrical Power and Energy Conference
  (EPEC)}.\hskip 1em plus 0.5em minus 0.4em\relax IEEE, 2021, pp. 125--130.

\bibitem{pan2022dynamic}
K.~Pan, P.~Palensky, and P.~M. Esfahani, ``Dynamic anomaly detection with
  high-fidelity simulators: A convex optimization approach,'' \emph{IEEE
  Transactions on Smart Grid}, vol.~13, no.~2, pp. 1500--1515, 2022.

\bibitem{bho2019on}
N.~Bhowmik, Y.~F.~A. Gaus, S.~Akçay, J.~W. Barker, and T.~P. Breckon, ``On the
  impact of object and sub-component level segmentation strategies for
  supervised anomaly detection within x-ray security imagery,'' in \emph{2019
  18th IEEE International Conference On Machine Learning And Applications
  (ICMLA)}, 2019, pp. 986--991.

\bibitem{lin2020anomaly}
T.-H. Lin and J.-R. Jiang, ``Anomaly detection with autoencoder and random
  forest,'' in \emph{2020 International Computer Symposium (ICS)}, 2020, pp.
  96--99.

\bibitem{gharaei2022}
R.~H. Gharaei, R.~Sharify, and H.~Nezamabadi-Pour, ``An efficient outlier
  detection method based on distance ratio of k-nearest neighbors,'' in
  \emph{2022 9th Iranian Joint Congress on Fuzzy and Intelligent Systems
  (CFIS)}, 2022, pp. 1--5.

\bibitem{zenati2018}
H.~Zenati, C.~S. Foo, B.~Lecouat, G.~Manek, and V.~R. Chandrasekhar,
  ``Efficient gan-based anomaly detection,'' 2018.

\bibitem{wu2019multiple}
D.~Wu, B.~Wang, D.~Precup, and B.~Boulet, ``Multiple kernel learning-based
  transfer regression for electric load forecasting,'' \emph{IEEE Transactions
  on Smart Grid}, vol.~11, no.~2, pp. 1183--1192, 2019.

\bibitem{wu2017boosting}
------, ``Boosting based multiple kernel learning and transfer regression for
  electricity load forecasting,'' in \emph{Joint European Conference on Machine
  Learning and Knowledge Discovery in Databases}.\hskip 1em plus 0.5em minus
  0.4em\relax Springer, 2017, pp. 39--51.

\bibitem{lin2021spatial}
W.~Lin, D.~Wu, and B.~Boulet, ``Spatial-temporal residential short-term load
  forecasting via graph neural networks,'' \emph{IEEE Transactions on Smart
  Grid}, vol.~12, no.~6, pp. 5373--5384, 2021.

\bibitem{lin2021residential}
W.~Lin, ``Residential electric load forecasting via attentive transfer of graph
  neural networks.''

\bibitem{wu2022efficient}
D.~Wu and W.~Lin, ``Efficient residential electric load forecasting via
  transfer learning and graph neural networks,'' \emph{IEEE Transactions on
  Smart Grid}, 2022.

\bibitem{fu2022reinforcement}
Y.~Fu, D.~Wu, and B.~Boulet, ``Reinforcement learning based dynamic model
  combination for time series forecasting,'' in \emph{Proceedings of the AAAI
  Conference on Artificial Intelligence}, vol.~36, no.~6, 2022, pp. 6639--6647.

\bibitem{dang2019q}
Q.~Dang, D.~Wu, and B.~Boulet, ``A q-learning based charging scheduling scheme
  for electric vehicles,'' in \emph{2019 IEEE Transportation Electrification
  Conference and Expo (ITEC)}.\hskip 1em plus 0.5em minus 0.4em\relax IEEE,
  2019, pp. 1--5.

\bibitem{wu2018optimizing}
D.~Wu, G.~Rabusseau, V.~Fran{\c{c}}ois-lavet, D.~Precup, and B.~Boulet,
  ``Optimizing home energy management and electric vehicle charging with
  reinforcement learning.''

\bibitem{wu2014neighborhood}
D.~Wu, H.~Zeng, and B.~Boulet, ``Neighborhood level network aware electric
  vehicle charging management with mixed control strategy,'' in \emph{2014 IEEE
  International Electric Vehicle Conference (IEVC)}.\hskip 1em plus 0.5em minus
  0.4em\relax IEEE, 2014, pp. 1--7.

\bibitem{dang2020ev}
Q.~Dang, D.~Wu, and B.~Boulet, ``Ev charging management with ann-based
  electricity price forecasting,'' in \emph{2020 IEEE Transportation
  Electrification Conference \& Expo (ITEC)}.\hskip 1em plus 0.5em minus
  0.4em\relax IEEE, 2020, pp. 626--630.

\bibitem{huang2020ensemble}
X.~Huang, D.~Wu, and B.~Boulet, ``Ensemble learning for charging load
  forecasting of electric vehicle charging stations,'' in \emph{2020 IEEE
  Electric Power and Energy Conference (EPEC)}.\hskip 1em plus 0.5em minus
  0.4em\relax IEEE, 2020, pp. 1--5.

\bibitem{dang2019advanced}
Q.~Dang, D.~Wu, and B.~Boulet, ``An advanced framework for electric vehicles
  interaction with distribution grids based on q-learning,'' in \emph{2019 IEEE
  Energy Conversion Congress and Exposition (ECCE)}.\hskip 1em plus 0.5em minus
  0.4em\relax IEEE, pp. 3491--3495.

\bibitem{xiong2015impact}
J.~Xiong, D.~Wu, H.~Zeng, S.~Liu, and X.~Wang, ``Impact assessment of electric
  vehicle charging on hydro ottawa distribution networks at neighborhood
  levels,'' in \emph{2015 IEEE 28th Canadian Conference on Electrical and
  Computer Engineering (CCECE)}.\hskip 1em plus 0.5em minus 0.4em\relax IEEE,
  2015, pp. 1072--1077.

\bibitem{Goodfellow2014generative}
I.~Goodfellow, J.~Pouget-Abadie, M.~Mirza, B.~Xu, D.~Warde-Farley, S.~Ozair,
  A.~Courville, and Y.~Bengio, ``Generative adversarial nets,'' in
  \emph{Advances in Neural Information Processing Systems}, Z.~Ghahramani,
  M.~Welling, C.~Cortes, N.~Lawrence, and K.~Weinberger, Eds., vol.~27.\hskip
  1em plus 0.5em minus 0.4em\relax Curran Associates, Inc., 2014.

\bibitem{jeong2021poster}
H.~Jeong, J.~Yu, and W.~Lee, ``Poster abstract: A semi-supervised approach for
  network intrusion detection using generative adversarial networks,'' in
  \emph{IEEE INFOCOM 2021 - IEEE Conference on Computer Communications
  Workshops (INFOCOM WKSHPS)}, 2021, pp. 1--2.

\bibitem{jiang2020discriminative}
T.~Jiang, W.~Xie, Y.~Li, and Q.~Du, ``Discriminative semi-supervised generative
  adversarial network for hyperspectral anomaly detection,'' in \emph{IGARSS
  2020 - 2020 IEEE International Geoscience and Remote Sensing Symposium},
  2020, pp. 2420--2423.

\bibitem{karuna2022generative}
E.~N. Karuna, P.~V. Sokolov, and D.~A. Gavrilic, ``Generative adversarial
  approach in natural language processing,'' in \emph{2022 XXV International
  Conference on Soft Computing and Measurements (SCM)}, 2022, pp. 111--114.

\bibitem{mudavathu2018aux}
K.~D.~B. Mudavathu, M.~V. P. C.~S. Rao, and K.~V. Ramana, ``Auxiliary
  conditional generative adversarial networks for image data set
  augmentation,'' in \emph{2018 3rd International Conference on Inventive
  Computation Technologies (ICICT)}, 2018, pp. 263--269.

\bibitem{hu2020generative}
Z.~Hu, T.~Turki, and J.~T.~L. Wang, ``Generative adversarial networks for
  stochastic video prediction with action control,'' \emph{IEEE Access},
  vol.~8, pp. 63\,336--63\,348, 2020.

\bibitem{akcay2019ganomaly}
S.~Akcay, A.~Atapour-Abarghouei, and T.~P. Breckon, ``Ganomaly: Semi-supervised
  anomaly detection via adversarial training,'' in \emph{Computer Vision --
  ACCV 2018}, C.~V. Jawahar, H.~Li, G.~Mori, and K.~Schindler, Eds.\hskip 1em
  plus 0.5em minus 0.4em\relax Springer International Publishing, 2019, pp.
  622--637.

\bibitem{Bhag2020study}
Bhagyashree, V.~Kushwaha, and G.~C. Nandi, ``Study of prevention of mode
  collapse in generative adversarial network (gan),'' in \emph{2020 IEEE 4th
  Conference on Information \& Communication Technology (CICT)}, 2020, pp.
  1--6.

\bibitem{koda2017on}
N.~Kodali, J.~Abernethy, J.~Hays, and Z.~Kira, ``On convergence and stability
  of gans,'' 2017.

\bibitem{tol2017adagan}
I.~O. Tolstikhin, S.~Gelly, O.~Bousquet, C.-J. SIMON-GABRIEL, and
  B.~Sch\"{o}lkopf, ``Adagan: Boosting generative models,'' in \emph{Advances
  in Neural Information Processing Systems}, I.~Guyon, U.~V. Luxburg,
  S.~Bengio, H.~Wallach, R.~Fergus, S.~Vishwanathan, and R.~Garnett, Eds.,
  vol.~30.\hskip 1em plus 0.5em minus 0.4em\relax Curran Associates, Inc.,
  2017.

\bibitem{Han_Chen_Liu_2021}
X.~Han, X.~Chen, and L.-P. Liu, ``Gan ensemble for anomaly detection,''
  \emph{Proceedings of the AAAI Conference on Artificial Intelligence},
  vol.~35, no.~5, pp. 4090--4097, 2021.

\bibitem{maher2019smart}
M.~M. M. Z.~A. Maher and S.~Sakr, ``{SmartML: A meta learning-based framework
  for automated selection and hyperparameter tuning for machine learning
  algorithms},'' in \emph{{EDBT: 22nd International Conference on Extending
  Database Technology}}, Lisbon, Portugal, Mar. 2019.

\bibitem{molina2012meta}
M.~M. Molina, J.~M. Luna, C.~Romero, and S.~Ventura, ``Meta-learning approach
  for automatic parameter tuning: A case study with educational datasets,'' in
  \emph{Proceedings of the 5th International Conference on Educational Data
  Mining}, ser. EDM 2012, Chania, Greece, 2012, pp. 180--183.

\bibitem{so2021exploring}
C.~So, ``Exploring meta learning: parameterizing the learning-to-learn process
  for image classification,'' in \emph{2021 International Conference on
  Artificial Intelligence in Information and Communication (ICAIIC)}, 2021, pp.
  199--202.

\bibitem{doke2021survey}
A.~Doke and M.~Gaikwad, ``Survey on automated machine learning (automl) and
  meta learning,'' in \emph{2021 12th International Conference on Computing
  Communication and Networking Technologies (ICCCNT)}, 2021, pp. 1--5.

\bibitem{lars2017auto}
L.~Kotthoff, C.~Thornton, H.~H. Hoos, F.~Hutter, and K.~Leyton-Brown,
  ``Auto-weka 2.0: Automatic model selection and hyperparameter optimization in
  weka,'' \emph{Journal of Machine Learning Research}, vol.~18, no.~25, pp.
  1--5, 2017.

\bibitem{feurer2015efficient}
M.~Feurer, A.~Klein, K.~Eggensperger, J.~Springenberg, M.~Blum, and F.~Hutter,
  ``Efficient and robust automated machine learning,'' in \emph{Advances in
  Neural Information Processing Systems}, C.~Cortes, N.~Lawrence, D.~Lee,
  M.~Sugiyama, and R.~Garnett, Eds., vol.~28.\hskip 1em plus 0.5em minus
  0.4em\relax Curran Associates, Inc., 2015.

\bibitem{provotar2019unsupervised}
O.~I. Provotar, Y.~M. Linder, and M.~M. Veres, ``Unsupervised anomaly detection
  in time series using lstm-based autoencoders,'' in \emph{2019 IEEE
  International Conference on Advanced Trends in Information Theory (ATIT)},
  2019, pp. 513--517.

\bibitem{schlegl2017unsupervised}
T.~Schlegl, P.~Seeb{\"o}ck, S.~M. Waldstein, U.~Schmidt-Erfurth, and G.~Langs,
  ``Unsupervised anomaly detection with generative adversarial networks to
  guide marker discovery,'' in \emph{Information Processing in Medical
  Imaging}, M.~Niethammer, M.~Styner, S.~Aylward, H.~Zhu, I.~Oguz, P.-T. Yap,
  and D.~Shen, Eds.\hskip 1em plus 0.5em minus 0.4em\relax Springer
  International Publishing, 2017, pp. 146--157.

\bibitem{AHMED201619}
M.~Ahmed, A.~{Naser Mahmood}, and J.~Hu, ``A survey of network anomaly
  detection techniques,'' \emph{Journal of Network and Computer Applications},
  vol.~60, pp. 19--31, 2016.

\bibitem{xiong2011group}
L.~Xiong, B.~P\'{o}czos, and J.~Schneider, ``Group anomaly detection using
  flexible genre models,'' in \emph{Advances in Neural Information Processing
  Systems}, J.~Shawe-Taylor, R.~Zemel, P.~Bartlett, F.~Pereira, and
  K.~Weinberger, Eds., vol.~24.\hskip 1em plus 0.5em minus 0.4em\relax Curran
  Associates, Inc., 2011.

\bibitem{chen2001one}
Y.~Chen, X.~S. Zhou, and T.~Huang, ``One-class svm for learning in image
  retrieval,'' in \emph{Proceedings 2001 International Conference on Image
  Processing (Cat. No.01CH37205)}, vol.~1, 2001, pp. 34--37 vol.1.

\bibitem{zimek2012survey}
A.~Zimek, E.~Schubert, and H.-P. Kriegel, ``A survey on unsupervised outlier
  detection in high-dimensional numerical data,'' \emph{Statistical Analysis
  and Data Mining}, vol.~5, no.~5, 2012.

\bibitem{zhou2017anomaly}
C.~Zhou and R.~C. Paffenroth, ``Anomaly detection with robust deep
  autoencoders,'' in \emph{Proceedings of the 23rd ACM SIGKDD International
  Conference on Knowledge Discovery and Data Mining}.\hskip 1em plus 0.5em
  minus 0.4em\relax New York, NY, USA: Association for Computing Machinery,
  2017, p. 665–674.

\bibitem{zong2018deep}
B.~Zong, Q.~Song, M.~R. Min, W.~Cheng, C.~Lumezanu, D.~Cho, and H.~Chen, ``Deep
  autoencoding gaussian mixture model for unsupervised anomaly detection,'' in
  \emph{International Conference on Learning Representations}, 2018.

\bibitem{Shen_Yu_Ma_Kwok_2021}
L.~Shen, Z.~Yu, Q.~Ma, and J.~T. Kwok, ``Time series anomaly detection with
  multiresolution ensemble decoding,'' \emph{Proceedings of the AAAI Conference
  on Artificial Intelligence}, vol.~35, no.~11, pp. 9567--9575, 2021.

\bibitem{chen2018auto}
Z.~Chen, C.~K. Yeo, B.~S. Lee, and C.~T. Lau, ``Autoencoder-based network
  anomaly detection,'' in \emph{2018 Wireless Telecommunications Symposium
  (WTS)}, 2018, pp. 1--5.

\bibitem{sakhnini2019}
J.~Sakhnini, H.~Karimipour, and A.~Dehghantanha, ``Smart grid cyber attacks
  detection using supervised learning and heuristic feature selection,'' in
  \emph{2019 IEEE 7th International Conference on Smart Energy Grid Engineering
  (SEGE)}, 2019, pp. 108--112.

\bibitem{zhu2020geometric}
Z.~Zhu, Z.~Wang, D.~Li, Y.~Zhu, and W.~Du, ``Geometric structural ensemble
  learning for imbalanced problems,'' \emph{IEEE Transactions on Cybernetics},
  vol.~50, no.~4, pp. 1617--1629, 2020.

\bibitem{han2020research}
Y.~Han, Y.~Ma, J.~Wang, and J.~Wang, ``Research on ensemble model of anomaly
  detection based on autoencoder,'' in \emph{2020 IEEE 20th International
  Conference on Software Quality, Reliability and Security (QRS)}, 2020, pp.
  414--417.

\bibitem{chen2018evo}
Z.~Chen, C.~K. Yeo, B.~S. Lee, C.~T. Lau, and Y.~Jin, ``Evolutionary
  multi-objective optimization based ensemble autoencoders for image outlier
  detection,'' \emph{Neurocomput.}, vol. 309, p. 192–200, 2018.

\bibitem{lichman2013UCI}
M.~Lichman, ``Uci machine learning repository,'' 2013.

\bibitem{borge2014ml}
R.~C. Borges~Hink, J.~M. Beaver, M.~A. Buckner, T.~Morris, U.~Adhikari, and
  S.~Pan, ``Machine learning for power system disturbance and cyber-attack
  discrimination,'' in \emph{2014 7th International Symposium on Resilient
  Control Systems (ISRCS)}, 2014, pp. 1--8.

\bibitem{bebis1994feed}
G.~Bebis and M.~Georgiopoulos, ``Feed-forward neural networks,'' \emph{IEEE
  Potentials}, vol.~13, no.~4, pp. 27--31, 1994.

\end{thebibliography}

%




\end{document}